%% file: root.tex
\documentclass[letterpaper, 10 pt, conference]{ieeeconf}  

\IEEEoverridecommandlockouts                              

\usepackage{graphicx}
\usepackage{amsmath}
\usepackage{amssymb}
\usepackage{booktabs}
\usepackage{bm}
\usepackage{soul}
\usepackage{multicol}
\usepackage{multirow}
\usepackage{amsmath}
\usepackage{tabularx}

\usepackage{color}

\newlength\savewidth\newcommand\shline{\noalign{\global\savewidth\arrayrulewidth
  \global\arrayrulewidth 1pt}\hline\noalign{\global\arrayrulewidth\savewidth}}

\usepackage{hyperref}
\hypersetup{pagebackref,breaklinks,colorlinks}
\usepackage{verbatim}
\usepackage[capitalize]{cleveref}

\crefname{section}{Sec.}{Secs.}
\Crefname{section}{Section}{Sections}
\Crefname{table}{Table}{Tables}
\crefname{table}{Tab.}{Tabs.}
\newcommand{\etal}{\textit{et al}. }
\newcommand{\ie}{\textit{i}.\textit{e}., }
\newcommand{\eg}{\textit{e}.\textit{g}. }
\usepackage{threeparttable}
\title{\LARGE \bf
SSP-Pose: Symmetry-Aware Shape Prior Deformation for Direct Category-Level Object Pose Estimation
}

\author{Ruida Zhang$^{1*}$, Yan Di$^{2*}$, Fabian Manhardt$^{3}$, Federico Tombari$^{2, 3}$, and Xiangyang Ji$^{4}$
\thanks{*Authors with equal contributions.}
\thanks{$^{1}$ Tsinghua University, zhangrd21@mails.tsinghua.edu.cn}%
\thanks{$^{2}$ Technical University of Munich.}%
\thanks{$^{3}$ Google}%
\thanks{$^{4}$ Tsinghua University}%
}

\begin{document}

\maketitle
\thispagestyle{empty}
\pagestyle{empty}

\begin{abstract}
Category-level pose estimation is a challenging problem due to intra-class shape variations. 
Recent methods deform pre-computed shape priors to map the observed point cloud into the normalized object coordinate space and then retrieve the pose via post-processing, \ie Umeyama's Algorithm. 
The shortcomings of this two-stage strategy lie in two aspects: 1) The surrogate supervision on the intermediate results can not directly guide the learning of pose, resulting in large pose error after post-processing. 2) The inference speed is limited by the post-processing step.
In this paper, to handle these shortcomings, we propose an end-to-end trainable network SSP-Pose for category-level pose estimation, which integrates shape priors into a direct pose regression network.
SSP-Pose stacks four individual branches on a shared feature extractor, where two branches are designed to deform and match the prior model with  the observed instance, and the other two branches are applied for directly regressing the totally 9 degrees-of-freedom pose and performing symmetry reconstruction and point-wise inlier mask prediction respectively.
Consistency loss terms are then naturally exploited to align the outputs of different branches and promote the performance.
During inference, only the direct pose regression branch is needed.
In this manner, SSP-Pose not only learns category-level pose-sensitive characteristics to boost performance but also keeps a real-time inference speed.
Moreover, we utilize the symmetry information of each category to guide the shape prior deformation, and propose a novel symmetry-aware loss to mitigate the matching ambiguity.
Extensive experiments on public datasets demonstrate that SSP-Pose produces superior performance compared with competitors with a real-time inference speed at about 25Hz. 
The codes will be released soon.

\end{abstract}

\input{sections/introduction}
\input{sections/related_work}
\input{sections/method}

\input{sections/experiment}

\input{sections/conclusion}





{\small
\bibliographystyle{IEEEtran}
\bibliography{egbib}
}

\end{document}

%% file: sections/introduction.tex
\section{INTRODUCTION}

Object pose estimation is a fundamental task in computer vision society, due to its wide applications in AR/VR~\cite{arvr}, robotics~\cite{robotics} and autonomous driving~\cite{driving1,driving2, gupnet}.
Researchers have made great progress in recent years and produced several reliable and efficient methods~\cite{labbe2020cosypose, GDRN, sopose}. 
However, most of them can only deal with a handful of instances and require per-object CAD models, which limits its real-world applications to some extent.
Recently, category-level pose estimation is proposed to predict full configurations of rotation, translation and 3D dimensions for previously unseen instances from a known set of categories.  
It's a much more challenging problem due to the lack of CAD models and the intra-class shape variations.

\begin{figure}[t]
  \centering
  \includegraphics[width=0.99\linewidth]{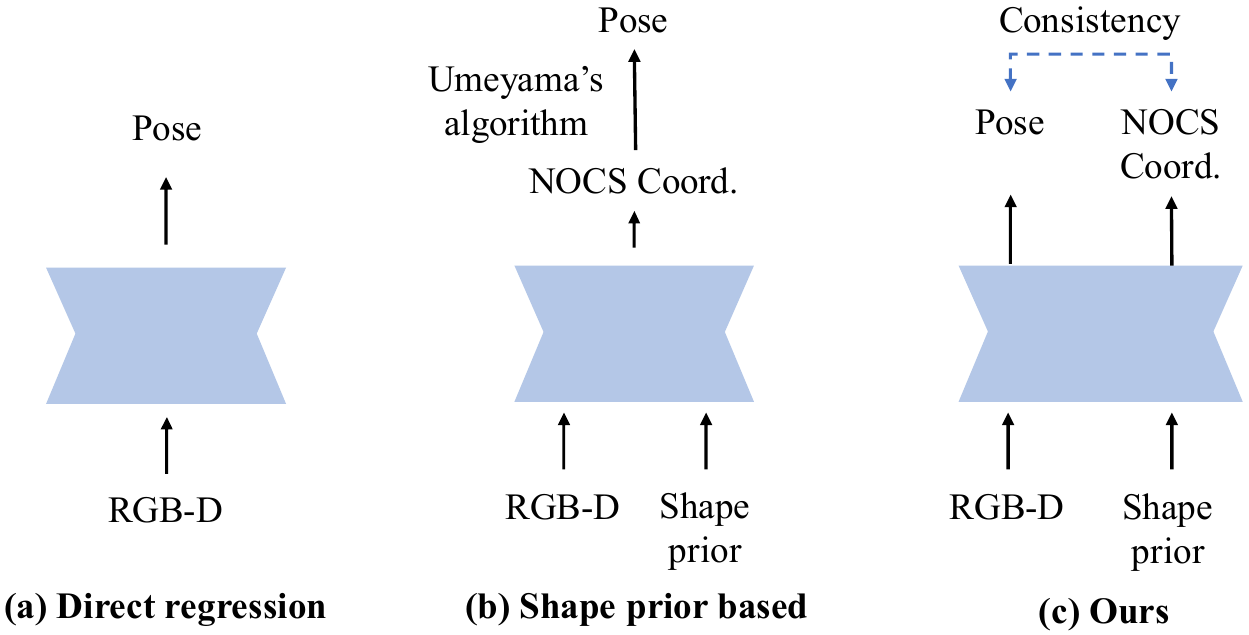}
  \caption{\textbf{Direct regression network \textit{vs} Shape prior-guided network \textit{vs} Ours.} Most shape prior guided methods are computational inefficient due to post-processing optimization for pose with Umeyama's algorithm~\cite{umeyama}, 
  while the direct regression methods suffer from the lack of prior shape knowledge.
  Our method takes advantage of both methods and leverages an auxiliary shape prior adaptation branch to aid the direct pose regression branch.
  }
  \label{teasor}
\end{figure}

Wang \etal~\cite{NOCS} propose the Normalized Object Coordinate Space (NOCS) and utilize a CNN network to establish correspondences between the observed point cloud and the NOCS coordinates.
Finally, the object pose is recovered with Umeyama’s algorithm~\cite{umeyama}. 
To tackle the intra-class variations, SPD~\cite{shape_deform} derives a point cloud based mean shape for each category and utilizes it as prior information.
It deforms the shape prior to the observed instance, and then maps the observed point cloud to the deformed shape to obtain the NOCS coordinates. 
SGPA~\cite{sgpa}, DO-Net~\cite{donet} and CR-Net~\cite{cr-net} follow SPD and utilize shape prior deformation to guide the learning. 
These methods show powerful strength in extracting shape-sensitive features, which further improves the NOCS coordinate prediction and the pose estimation. 
However, this strategy suffers from two shortcomings.
First, the accuracy of pose estimation is expected to monotonically improve with the quality of intermediate results, which is proved to be infeasible in~\cite{GDRN}.
The error on the intermediate results does not necessarily reflect the actual error of pose, especially for symmetric objects.
Moreover, the inference speed is limited by the time-consuming post-processing step, \ie Umeyama’s algorithm~\cite{umeyama}.

Meanwhile, another line of methods that predict the pose directly has aroused wide research interest.
FS-Net~\cite{fs-net} leverages disentangled representation of rotation to tackle ambiguity caused by symmetry and proposes the 3D bounding box deformation based data augmentation. 
DualPoseNet~\cite{dualposenet} employs an explicit decoder to regress the pose directly and an implicit decoder to reconstruct the object shape.
Direct methods don't depend on the coordinate mapping process in comparison with two-stage methods, which makes them inherently faster and end-to-end trainable.
However, they suffer from the lack of category-level prior knowledge and typically their performance is inferior to shape prior guided competitors. 

In this paper, to handle these shortcomings, we propose SSP-Pose, an end-to-end trainable network for category-level pose estimation, which takes advantage of both prior-guided methods and direct regression methods for high accuracy and efficiency.
Given an RGB-D image, SSP-Pose first detects the target object and then extracts point-wise features from the point cloud back-projected from the corresponding depth map of the object as well as the category shape prior.
Four sub-branches are attached to the features to perform individual prediction tasks.
The first two branches predict auxiliary information that deforms and associates the shape prior to the observed instance so that each observed point is connected with its corresponding point in NOCS.
The third branch, following FS-Net~\cite{yangpvpose}, directly regresses all 9 degrees-of-freedom (DoF) pose parameters in the disentangled representation.
The fourth branch performs symmetry reconstruction to jointly reconstruct the object shape utilizing symmetry prior and implicitly guide the learning of pose.
Moreover, the fourth branch predicts a point-wise inlier mask to distinguish outliers and improve the robustness of our method. 
A geometry-guided consistency loss term is naturally introduced to align the outputs from different branches and boost the performance.
Note that during inference, only the pose regression branch is needed thus the inference speed is guaranteed to be efficient.
Meanwhile, we leverage the symmetric characteristics of several categories \eg \textit{bottle, can, laptop} to guide the learning of shape prior deformation and introduce a symmetry-aware coordinate loss to mitigate the matching ambiguity in the matching step.

In summary, our main contributions are as follows:
\begin{itemize}
\setlength{\itemsep}{0pt}
\setlength{\parsep}{0pt}
\setlength{\parskip}{0pt}
\item 
We propose an end-to-end category-level pose estimation network SSP-Pose, which deforms and associates the shape prior model to the observed object to establish consistency with the directly regressed pose, boosting performance whilst keeping a fast inference speed.
\item 
We utilize the symmetric prior of some categories, \eg \textit{bottle, laptop}, to guide shape prior deformation. 
Moreover, we propose a symmetry-aware matching loss to mitigate the ambiguity in the shape prior matching step, stabilizing and promoting the learning of category-level pose-sensitive features.
\item 
SSP-Pose produces state-of-the-art performance on common public benchmarks, whilst achieving real-time inference speed at about 25 FPS.
\end{itemize}

%% file: sections/related_work.tex
\section{RELATED WORKS}
\subsection{Instance-level Object Pose Estimation}
In instance-level object pose estimation, the network is trained and tested on several known objects with CAD models.
Recently, many efforts have been made to estimate 6DoF pose from RGB images or RGB-D data.
As for RGB-only methods, end-to-end methods \cite{liu2016ssd, Kehl2017, xiang2017posecnn, manhardt2018deep, manhardt2019explaining, li2019deepim, labbe2020cosypose} regress the pose parameters directly, while two-stage methods~\cite{hybridpose,peng2019pvnet,zakharov2019dpod,hodan2020epos,park2019pix2pose} first establish 2D-3D correspondences by predicting the 3D coordinate for each pixel or detecting pre-defined keypoints, and then utilize P\textit{n}P/RANSAC algorithm to solve the pose from intermediate results. 
Another line of methods\cite{li2019cdpn, GDRN,hu2020single,sopose} jointly take advantage of direct pose regression and correspondence-based pose recovery to boost performance. 
As for methods based on RGB-D data, FFB6D~\cite{FFB6D} and DenseFusion~\cite{densefusion} combine appearance and geometry information by iterative dense feature fusion, while PVN3D~\cite{pvn3d} proposes a 3D voting scheme to detect keypoints. 
Instance-level methods are limited in practice since they can deal with merely a handful of objects and require precise CAD models.

\subsection{Category-level Object Pose Estimation}
Category-level pose estimation aims to predict the 6DoF pose along with the 3D metric size of previously unseen objects from a known set of categories. 
Recently, Wang~\etal~\cite{NOCS} proposes the \textit{Normalized Object Coordinate Space} (NOCS). 
They represent the observed point cloud with NOCS and recover the pose by the Umeyama's algorithm~\cite{umeyama}.
CASS~\cite{cass} learns a canonical shape space by VAE~\cite{vae} to obtain a view-factorized RGB-D embedding. 
CenterSnap~\cite{centersnap} presents a one-stage pipeline to reduce the computational cost.
6-PACK~\cite{6dpack} recovers the pose by tracking inter-frame motion of the object. 
CAPTRA~\cite{captra} tracks the object pose by canonicalizing the observed point cloud using the previous predictions, while BundleTrack~\cite{bundletrack} employs pose graph optimization to enforce spatiotemporal consistency. 

To tackle the intra-class shape variation, which is considered as the main bottleneck in category-level pose estimation~\cite{cass,shape_deform}, shape priors are widely utilized~\cite{shape_deform, sgpa, donet}. 
Most of them first deform the shape prior to the observed instance and match the observed point cloud with the deformed prior model to retrieve the NOCS coordinates. 
Then Umeyama's algorithm is adopted to recover the pose from the NOCS coordinates.
Although they have achieved large improvements on several benchmarks, the performance is still far from excellent.

Meanwhile, other methods directly regress the pose parameters. FS-Net~\cite{fs-net} proposes a decoupled mechanism for rotation,  
while DualPoseNet~\cite{dualposenet} employs an explicit decoder to regress the pose directly and an implicit decoder to reconstruct the object shape. 
It refines the pose estimation by enforcing consistency between the two decoders.
However, they are unable to incorporate prior information.
In this paper, we integrate shape priors into the direct regression network, which ensures efficient and accurate inference.

\begin{figure*}[t]
  \centering
  ~\\[3pt]
  \includegraphics[width=0.99\linewidth]{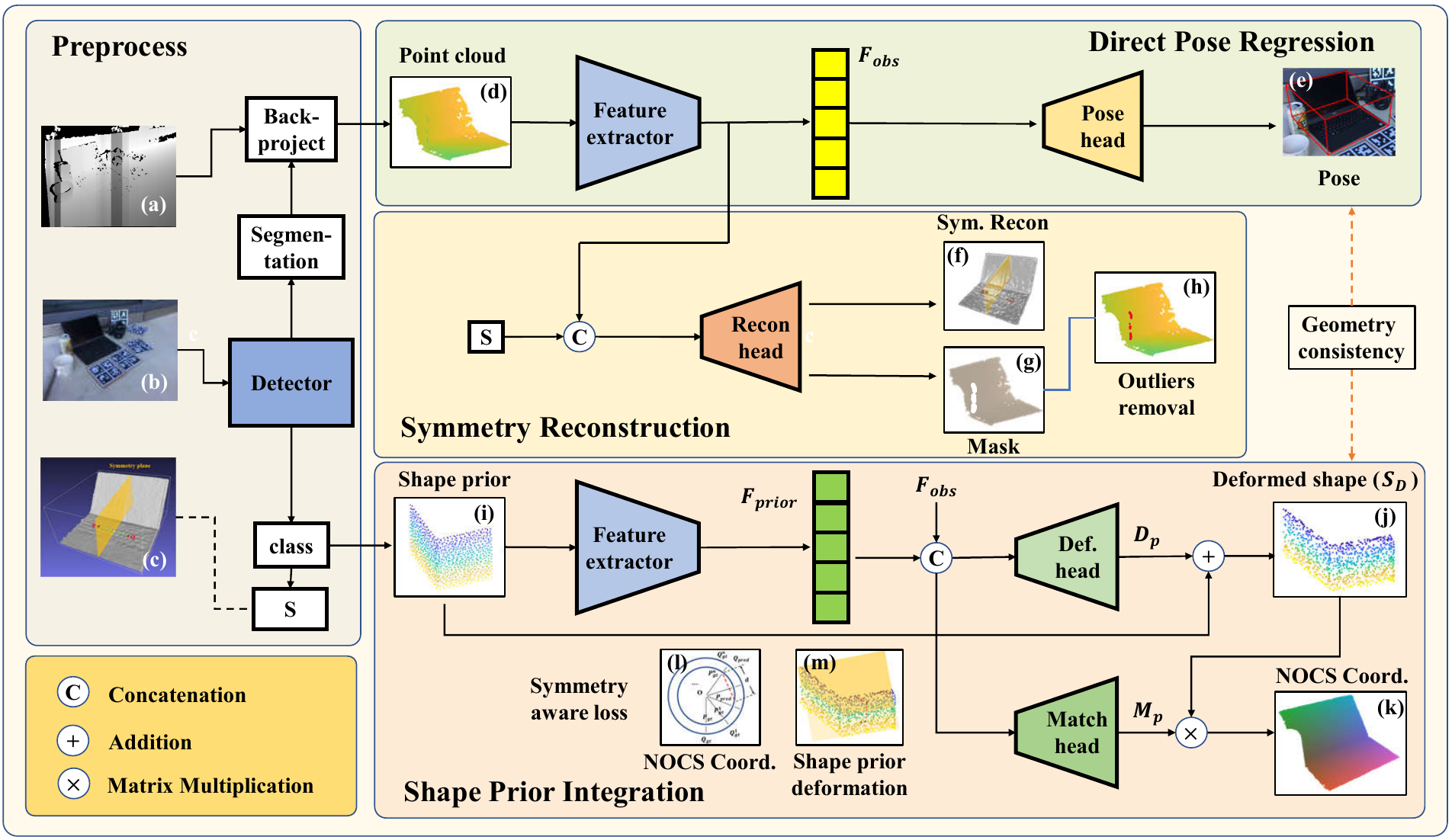}
  \caption{\textbf{Overview of SSP-Pose.} 
  We first use an off-the-shelf object detector (\eg MaskRCNN~\cite{maskrcnn}) to segment out the object of interest (b) and generate the back-projected point cloud 
  from the depth map (a).
  Then we adopt 3DGCN(\cite{3DGC} as the encoder to extract features of the observed point cloud $\mathcal{F}_{obs}$.
  The pose regression head takes $\mathcal{F}_{obs}$ as input and predicts the pose $R, t, s$ (e).
  We feed $\mathcal{F}_{obs}$ and the symmetry type $S$ (c) of the category into the reconstruction head (Recon. head). 
  It performs symmetry reconstruction (f) and predicts point-wise mask for outlier removal (g, h).
  Meanwhile, we use the same encoder to extract the feature of the mean shape (i) $\mathcal{F}_{prior}$. 
  The deformation head (Def. head) and matching head take both $\mathcal{F}_{obs}$ and $\mathcal{F}_{prior}$ as input. 
  The deformation head deforms the mean shape to the observed instance (j), and the matching head matches each observed point with the deformed model and obtains the NOCS coordinates (k). 
  We design a geometry-guided consistency term between the predicted pose and the NOCS coordinates to boost performance.
  Moreover, we propose a symmetry-aware loss to eliminate ambiguity for coordinate prediction and shape deformation (l, m).
  }
  \label{pipeline}
\end{figure*}

%% file: sections/method.tex
\section{SSP-Pose}

A schematic overview of SSP-Pose is presented in Fig~\ref{pipeline}. 
We first segment the region of objects of interest using an off-the-shelf object detector (e.g. MaskRCNN~\cite{maskrcnn}). 
We randomly sample 1024 points from the back-projected point cloud and feed them into our network. 
A 3D graph convolutional network (3DGC~\cite{3DGC}) based feature extractor is applied to extract point-wise features from the observed point cloud and shape prior respectively.
Then four parallel branches are attached for independent tasks.
The deformation head and matching head are designed to deform and associate the prior model with the observed instance.
The pose regression head is adopted for direct regression of the pose parameters. 
And the reconstruction head performs symmetry reconstruction as well as point-wise inlier mask prediction.
Geometry-guided consistency terms among these branches are finally enforced to align the outputs and boost the performance.

\subsection{Network Architecture \label{sec:spd}}
\textbf{Feature Extractor.}
Inspired by FS-Net~\cite{fs-net}, we choose 3DGC~\cite{3DGC}, which is insensitive to shift and scale, to extract point-wise features.
Specifically, we adopt the same 3DGC network to extract shape-sensitive features $\mathcal{F}_{obs}$, $\mathcal{F}_{prior}$ from both the observed point cloud $P_{obs}$ and the shape prior $P_{prior}$, allowing for shape prior guided shape reconstruction and pose estimation. 

\textbf{Pose regression head.}
Previous methods typically harness RANSAC-based Umeyama's algorithm~\cite{umeyama} to filter outliers and recover the pose.
However, RANSAC is vulnerable to systematic erroneous point matches, \eg the mistakenly established point matches caused by symmetry, which leads to performance deterioration.
Thus we design an end-to-end pose regression head and adopt three other auxiliary heads for cross-task consistency.
We feed $\mathcal{F}_{obs}$ to our pose head and regress the 9DoF pose directly, ensuring fast inference speed and robust pose estimation.  
We follow FS-Net~\cite{fs-net} to parameterize the translation and size with the residual representation, and parameterize the rotation in a geometry-aware manner. 
Specifically, for translation $\bm{t}$, given the predicted residual translation $\bm{t_r}$ and the mean of the observed point cloud $P_M$, we recover $\bm{t} = \bm{t_r} + P_M$.
Similarly, given the estimated residual size $\bm{s_r}$ and the pre-computed category mean size $S_M$, we obtain the scale $\bm{s} = \bm{s_r} + S_M$, where $\bm{s}=\{s_x, s_y, s_z\}$.
For rotation $\bm{R}$, we predict two rotation vectors along $x$ and $y$ axis $r_x, r_y$ respectively, which are also the first two columns of the rotation matrix.
For rotational symmetric objects which are symmetric around an axis, $r_x$ contains ambiguity and only $r_y$ is used to solve the rotation information, as shown in Fig.~\ref{symmetry}. 

\textbf{Reconstruction head.}
We adopt a reconstruction head for symmetry-aware reconstruction and outlier removal, which are introduced in Sec.~\ref{sec:sym},~\ref{sec:filter} respectively.
The reconstruction head takes $\mathcal{F}_{obs}$ and the category-level symmetry type $S$ as input, and outputs the mirrored input points (Sec.~\ref{sec:sym}) and the point-wise inlier mask (Sec.~\ref{sec:filter}). 

\textbf{Deformation and matching heads.}
These two heads take both $F_{obs}$ and $F_{prior}$ as input, and output the deformation field $D_{p}$ and the matching matrix $M_{p}$ respectively. 
$D_{p}$ deforms the shape prior model to the observed instance, and $M_{p}$ matches each observed point with the points in the deformed model.
Similar to SPD~\cite{shape_deform}, the NOCS coordinate $C$ of the target object is estimated as follows,
\begin{equation}
    C = M_{p}(P_{prior} + D_{p})
    \label{eq:prior}
\end{equation}

\subsection{Geometry-Guided Consistency \label{sec:con}}
We exploit the underlying geometry consistency between the predicted NOCS coordinate $C$ and the predicted pose, and transform it into a loss term to align both outputs and further improve the performance. 
Given the predicted pose $\bm{R}, \bm{t}, \bm{s}$, the NOCS coordinate of the point $p$ should be 
\begin{equation}
c_{p} = R^T (p - t) / L
    \label{eq:pt-coor}
\end{equation}
where $L$ is the diagonal length of the bounding box.

Thus our consistency loss term is defined as follows,
\begin{equation}
\mathcal{L}_{con} = \left| c_{p} - \overline{c} \right|
    \label{eq:consistency-loss}
\end{equation}
where $\overline{c}$ is the predicted NOCS coordinate. 

\begin{figure}[t]
  \centering
  \includegraphics[width=0.9\linewidth]{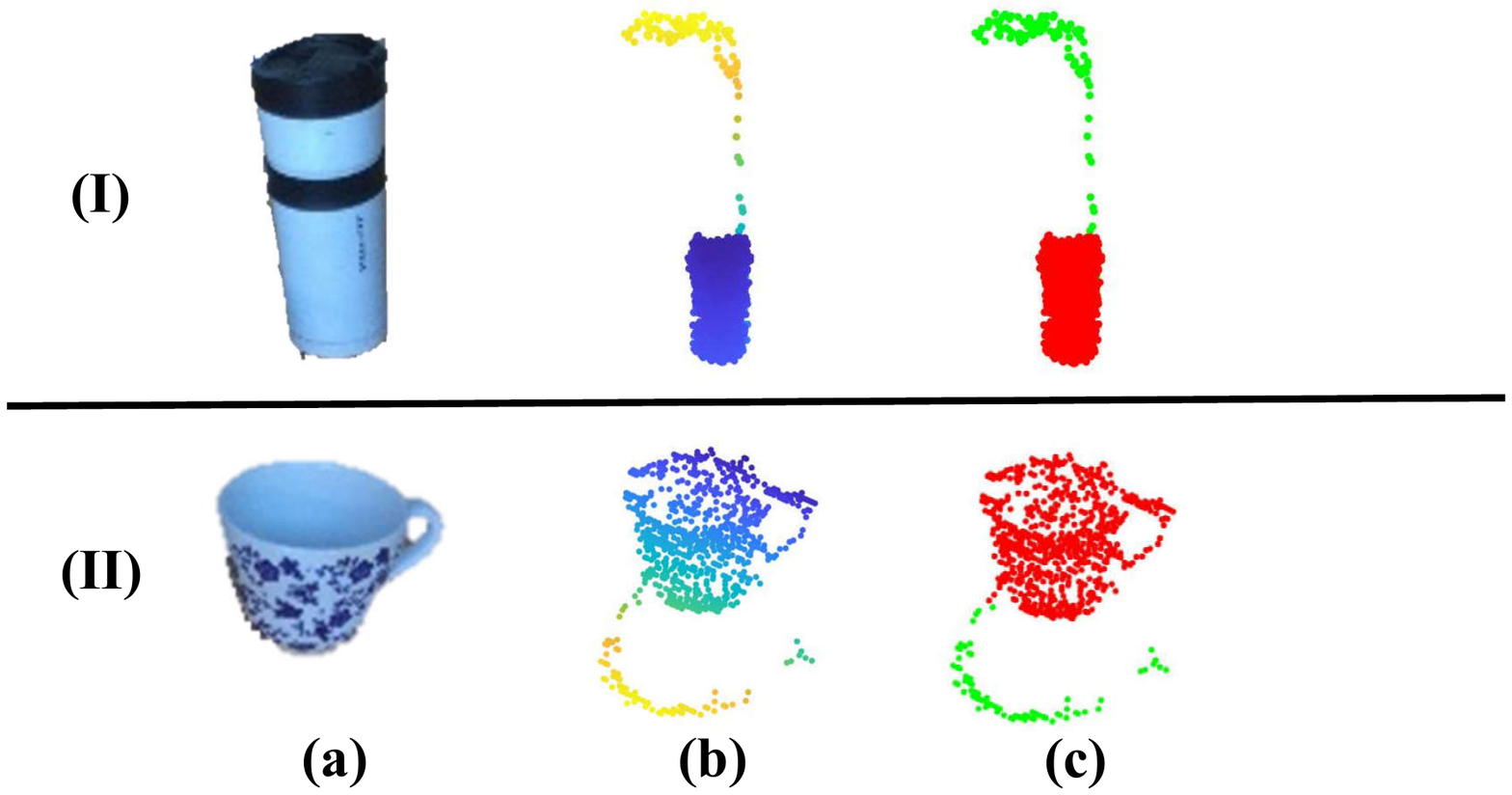}
  \caption{\textbf{Illustration of outlier removal.}
  \textbf{(I)},~\textbf{(II)} are two examples for outlier removal.
  Using the segmentation results from RGB images \textbf{(a)}, we back-project the depth map to retrieve point cloud \textbf{(b)}. Due to the limitation of depth sensor, the point cloud contains a non-negligible number of outliers. 
  We compute the ground truth location $p_{gt}$ for each point $p$ using the ground truth NOCS coordinate (red points in \textbf{(c)}) and compare $p_{gt}$ with the observed location $p_{obs}$ to detect outliers (green points in \textbf{(c)}).
  }
  \label{fig:filter}
\end{figure}

\subsection{Outlier Removal \label{sec:filter}}
We use an off-the-shelf object detector (\eg MaskRCNN~\cite{maskrcnn}) to segment the object of interest and back-project the corresponding depth map to obtain the point cloud as the input for the network.
Point cloud generated in this way contains a non-negligible number of outliers (as shown in Fig.~\ref{fig:filter}~\textbf{(b)}), which may mislead the network in pose estimation. 
In order to enhance the robustness of the training process, an outlier removal process is needed. 

In Fig.~\ref{fig:filter}, we demonstrate our outlier removal process.
We use the ground truth NOCS coordinate and pose to calculate the ground truth location $p_{gt}$ for each observed point $p$  (red points in Fig.~\ref{fig:filter}~\textbf{(c)}). 
\begin{equation}
p_{gt} = R(Lc)   + t
    \label{gt-location}
\end{equation}
where $L=\sqrt{s^2_x+s^2_y+s^2_z}$ stands for the diagonal length of the object bounding box.
We detect the outliers by comparing $p_{gt}$ with $p$ (green points in Fig.~\ref{fig:filter}~\textbf{(c)}).
Specifically, the point is considered to be an outlier if $|p-p_{gt}|>\lambda_{pt}$, where $\lambda_{pt}$ is the distance threshold. 
We set $\lambda_{pt}=0.1$ in this paper.

The reconstruction head predicts a point-wise mask indicating whether a point is an inlier, which guides the network to identify outliers and improves the robustness of our method.
Please note that we do not adopt outlier removal as a pre-processing step like in FS-Net~\cite{fs-net}.
The advantages lie in two points.
First, we enforce the network to distinguish outliers out of the point cloud itself, enabling robust performance when the outlier removal process fails in inference.
Second, only the simple direct pose regression network is needed during inference, guaranteeing fast running speed. 

\subsection{Symmetry Analysis \label{sec:sym}}
\begin{figure}[t]
  \centering
  ~\\[3pt]
  \includegraphics[width=0.9\linewidth]{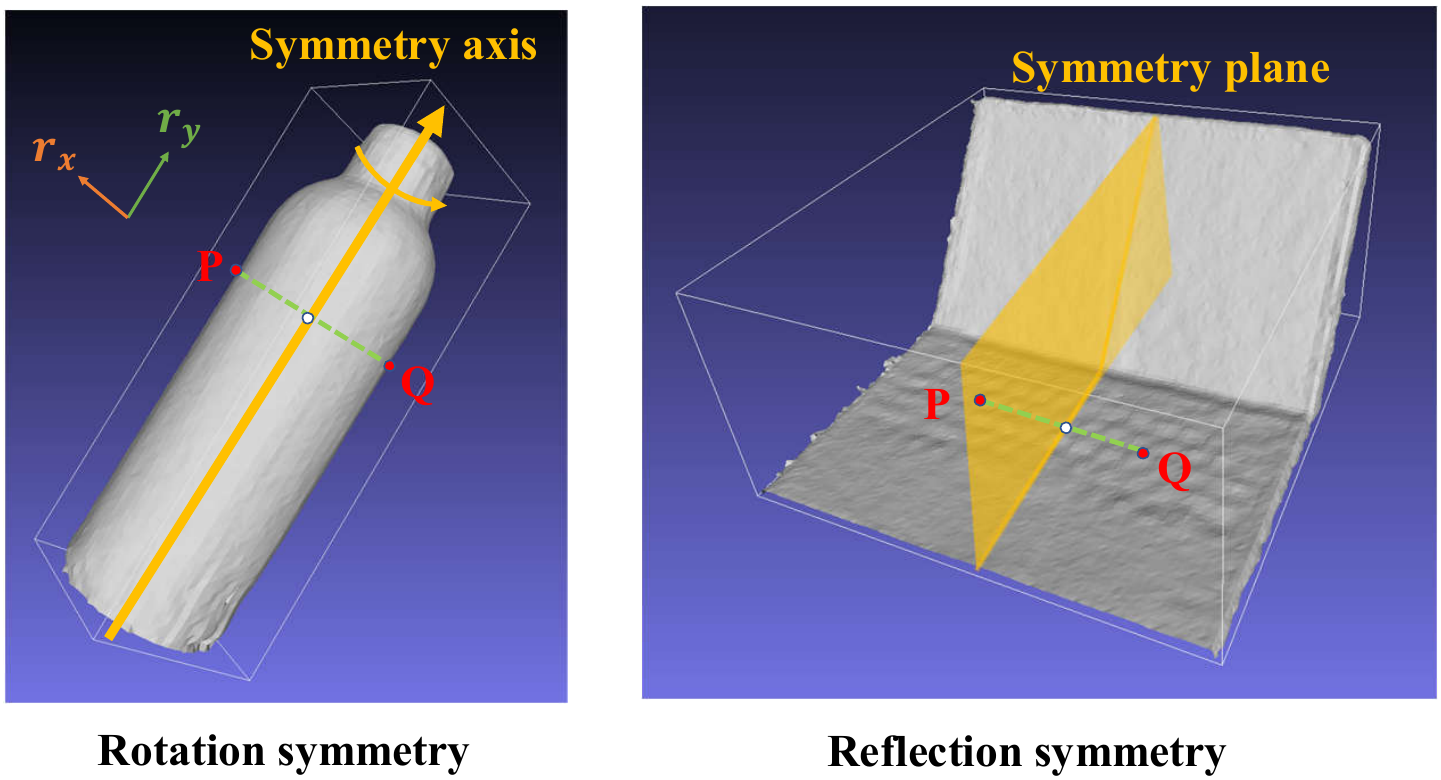}
  \caption{\textbf{Demonstration of two types of symmetry.}  Objects with \textbf{rotational symmetry} is symmetric along an axis, while objects with \textbf{reflection symmetry} is symmetric \textit{w.r.t.} a plane. Point \textbf{Q} is the mirrored point of point \textbf{P}.
  }
  \label{symmetry}
\end{figure}

We focus on two common types of symmetry: \textbf{Reflection Symmetry} and \textbf{Rotational Symmetry} \cite{donet}, which is demonstrated in Fig.~\ref{symmetry}.
Objects with {rotational symmetry} is symmetric along an axis, while objects with {reflection symmetry} is symmetric \textit{w.r.t.} a plane.
To simplify the illustration, we assume that the symmetry plane for reflection symmetry is on the $xy$-plane, the symmetry axis is $y$ axis for rotational symmetry.
For arbitrary category, we can adjust the canonical pose to guarantee the assumption. 

\begin{figure}[t]
  \centering
  \includegraphics[width=0.9\linewidth]{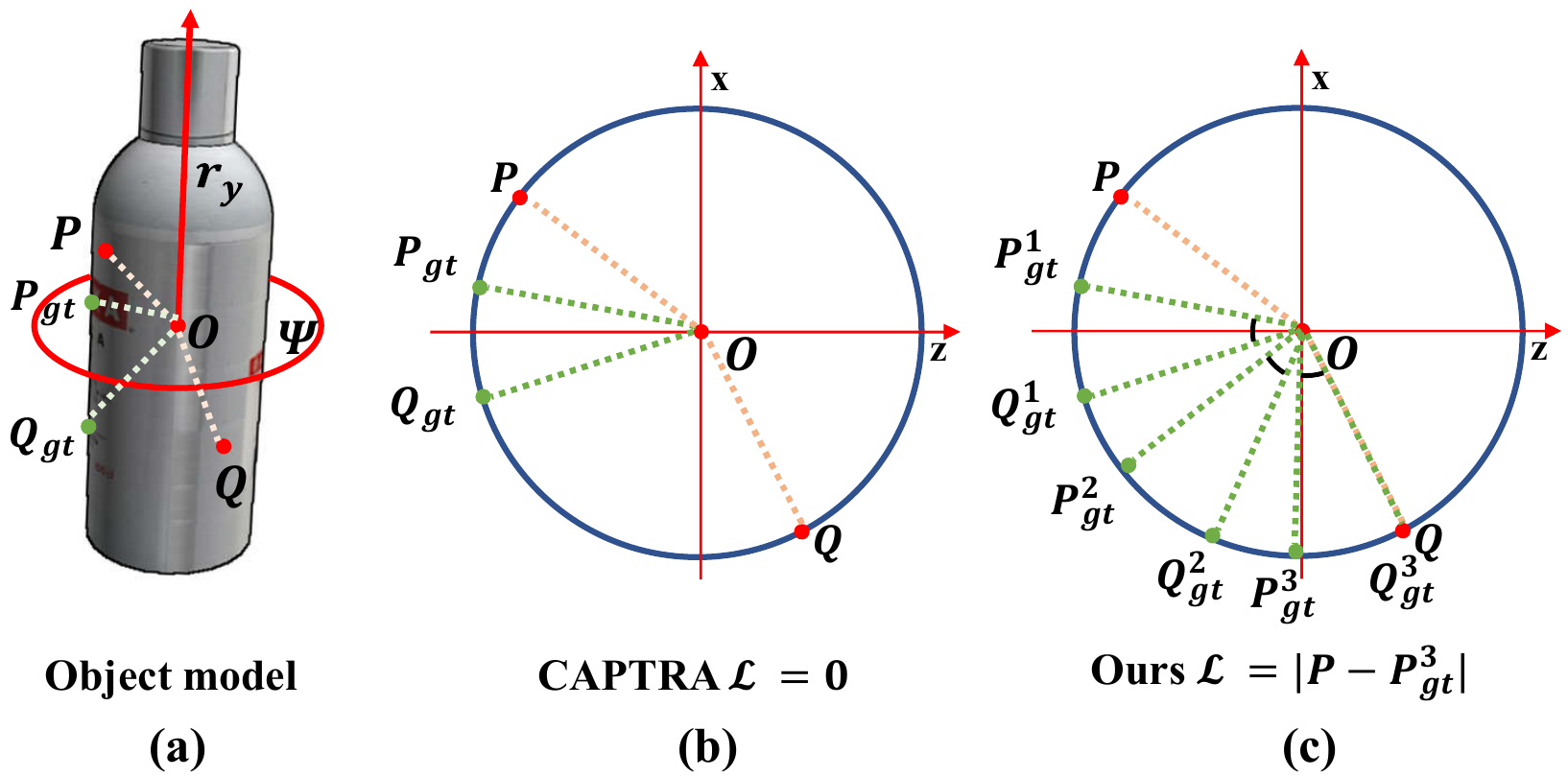}
  \caption{\textbf{Demonstration of symmetry-aware coordinate loss.}  
  We take the symmetric category, \ie \textit{bottle} as an example (a).
  Since the $y$ axis in the object coordinate coincides with the symmetry axis $r_y$, we only need to consider the $x, z$ axis.
  Thus we project the predicted NOCS coordinates on the plane $\Psi$ to simplify the illustration.
  In (b) and (c), $P$ and $Q$ are the predicted NOCS coordinates of two observed points and $P_{gt}, Q_{gt}$ are the corresponding ground truth.
  \textbf{O} denotes the projection of $r_y$. 
  Since CAPTRA\cite{captra} only considers the radius distance, as shown in (b), although the loss is zero, the predicted NOCS coordinates are still not fully constrained.
  Unlike CAPTRA, in (c), we first generate a set of candidate rotation matrix $R^i_{gt}, i=1,...,n$ by uniformly sampling on the circle.
  Then for each candidate pose $\{R^i_{gt}, t_{gt}\}$, we generate the candidate NOCS coordinates $N_c$ for all observed points, and compute the loss as in Eq.~\ref{loss-nocs}. 
  This loss alleviates the ambiguity caused by symmetry and guides the network to learn symmetry-aware geometric characteristics of each category.
  }
  \label{corr-loss}
\end{figure}

\textbf{Symmetry reconstruction.}
Previous methods (\cite{fs-net, cass}) adopt the point cloud reconstruction as an auxiliary task to improve the generalization ability of the network.
Inspired by DO-Net~\cite{donet}, we reconstruct the mirrored point for each observed point according to the symmetry type of the category (Figure~\ref{symmetry}). 
For category with no symmetry,  \eg \textit{camera}, we directly reconstruct the input point cloud. 
For category with rotational symmetry, \eg \textit{bowl, bottle, can}, the mirrored point is the corresponding point of the observed point \textit{w.r.t.} the symmetry axis.
For category with only reflection symmetry, \eg \textit{laptop}, the mirrored point can be obtained by flipping \textit{w.r.t.} the reflection plane.
In summary, for a point $p$, its corresponding mirrored point $\mathcal{F}_{mir}(p)$ under the canonical view is defined as follows, 
\begin{equation}
\mathcal{F}_{mir}(p) = \begin{cases}
[p_x, p_y, p_z] &  no \ symmetry \\
[-p_x, p_y, -p_z] &  rotational \ symmetry \\
[p_x, p_y, -p_z] &  reflection \ symmetry \\
\end{cases}
\end{equation}

\noindent The loss is defined as
\begin{equation}
\mathcal{L}_{recon} = \mathop{avg}_{p \in \mathcal{P}} \left | R(\mathcal{F}_{mir}(R^{-1}(p-t))) + t - \overline{p} \right |
    \label{eq:sym-recon-loss}
\end{equation}
where $\mathcal{P}$ is the input point cloud and $\overline{p}$ is the predicted mirrored point of $p$.
In this way, the network not only reconstructs the observed point cloud, but also implicitly learns the symmetry-aware geometric characteristics for pose estimation.

\textbf{Symmetry-aware shape deformation loss.}
We propose a symmetry-aware shape deformation loss   $\mathcal{L}^{SP}_{sym}$ in order to harness symmetry as prior information for shape deformation  (Fig.~\ref{pipeline} (j)). 
Since the category symmetry type remains unchanged after deformation, the mirrored points should also lie on the surface of the object.
We employ the Chamfer Distance between the mirrored point cloud $\mathcal{S}_{M}$ and the deformed shape $\mathcal{S}_D$ as the loss function. 
Specifically, the symmetry-aware shape recovery loss is defined as follows
\begin{equation}
 \mathcal{L}^{SP}_{sym} = \mathop{avg}_{p \in \mathcal{S}_D} \min_{q \in \mathcal{S}_{M}} \left|p - q \right| +  \mathop{avg}_{q \in \mathcal{S}_M} \min_{p \in \mathcal{S}_D} \left|p - q \right|
\label{loss-sym-shape}
\end{equation}

\textbf{Symmetry-aware coordinate loss.}
The pose of objects with rotational symmetry (or with more than one reflection planes) is ambiguous since multiple candidate poses can be obtained by rotating it around the symmetry axis.
Therefore, the pose-dependent NOCS coordinates (Fig.~\ref{pipeline} (k)) also contain ambiguity.
Forcing the network to predict the coordinates of symmetric objects under a specific pose will confuse the network.
To address this issue, CAPTRA\cite{captra} utilizes $\sqrt{c_x^2+c_z^2}$ for supervision, where NOCS coordinate $c=\{c_x, c_y, c_z\}$. 
Since the $y$ axis coincides with the symmetric axis $r_y$, $c_y$ can be omitted.
However, such a supervision manner may fail in several cases.
In Fig~\ref{corr-loss}, we provide an example to explain where the technique fails in detail. 

In order to tackle this problem, we propose a symmetry-aware NOCS coordinates loss, which alleviates the ambiguity caused by symmetry and guides the predicted coordinates to correspond to a unique pose.
We first uniformly sample a set of candidate rotation matrix $R^{i}_{gt}, i=1,...,n$ according to the symmetry type, as shown in Fig.~\ref{corr-loss} (c).
Then for each candidate pose $\{R^i_{gt}, t_{gt}\}$, we generate the candidate NOCS coordinates $N_c$ for all points, and compute the loss as the minimum $L_1$ distance between the predicted NOCS coordinates $C$ and all the candidates in $N_c$,
\begin{equation}
\mathcal{L}_{coor}^{SP} = \min_{C^{*} \in N_c}\frac{1}{|C|}\sum _{p \in C, q\in C^{*}}|p-q|
\label{loss-nocs}
\end{equation}
where $|C|$ denotes the point number in $C$.

\input{table/tab_real}
\input{table/tab_ablation}

\subsection{Overall training objective \label{sec:loss}}
The overall training objective is defined as follows,

\begin{equation}
\mathcal{L} = \lambda_{1}\mathcal{L}_{pose} +  \lambda_{2}\mathcal{L}_{SP}  + \lambda_{3}\mathcal{L}_{mask} + \lambda_{4}\mathcal{L}_{recon} + \lambda_{5}\mathcal{L}_{con}
\label{loss-all}
\end{equation}
For $\mathcal{L}_{pose}$, we follow FS-Net~\cite{fs-net} to supervise $\bm{R}, \bm{t}$ and $\bm{s}$ with the ground truth. 
We replace the L2 loss with L1 loss for faster convergence.
$\mathcal{L}_{SP}$ includes the loss terms introduced in Equation~\ref{loss-sym-shape},~\ref{loss-nocs}, along with loss terms for $D_p$ and $M_p$ from SPD~\cite{shape_deform}. 
We use L1 loss function for $\mathcal{L}_{mask}$ to supervise the point-wise mask prediction.
$\mathcal{L}_{recon}$ and $\mathcal{L}_{con}$ is introduced in Eq.~\ref{eq:consistency-loss} and \ref{eq:sym-recon-loss}.
Please note that we only calculate the loss terms  $\mathcal{L}_{SP}$, $\mathcal{L}_{con}$ and $\mathcal{L}_{recon}$ on the inliers.

%% file: table/tab_real.tex
\begin{table*}[t]
\centering
~\\[-5pt]
\caption{{Comparison with state-of-the-art methods on NOCS-REAL275 dataset.}
}
\begin{threeparttable}

\begin{tabular}{c|cc|ccc|cccc|c}
\shline
Method & Direct & Prior & $IoU_{25}$ & $IoU_{50}$ & $IoU_{75}$ &  $5^{\circ}2cm$& $5^{\circ}5cm$& $10^{\circ}5cm$& $10^{\circ}10cm$ & Speed(FPS)\\
\hline
NOCS~\cite{NOCS} & & & \textbf{84.9} & 80.5 & 30.1 & 7.2 & 10.0 & 25.2 & 26.7 & 5 \\
CASS~\cite{cass} & & & 84.2 & 77.7 & - & - & 23.5 & 58.0 & 58.3 & - \\
SPD~\cite{shape_deform} & &\checkmark & 83.4 & 77.3 & 53.2 & 19.3 & 21.4 & 54.1 & - & 4 \\
CR-Net~\cite{cr-net} & & \checkmark& - & 79.3 & 55.9 & 27.8 & 34.3 & 60.8 & - & - \\
SGPA~\cite{sgpa}  & &\checkmark& - & 80.1 & 61.9 & \textbf{35.9} & {39.6} & {70.7} & - & - \\
DO-Net~\cite{donet} & &\checkmark& - & 80.4 & {63.7} & 24.1 & 34.8 & 67.4 & - & 10 \\
\hline
DualPoseNet~\cite{dualposenet} &\checkmark& & - & 79.8 & 62.2  & 29.3 & 35.9 & 66.8 & - & 2 \\
FS-Net~\cite{fs-net} & \checkmark& & -& - & - & - & 28.2 & 60.8 & 64.6 & 20
\\
FS-Net(Ours) & \checkmark& & $84.0$& {81.1} & $52.0$ & $19.9$ & 33.9 & 69.1 & 71.0 & 20 \\
\hline
Ours & \checkmark&\checkmark & {84.0} &  \textbf{82.3} &  \textbf{66.3} & {34.7} & \textbf{44.6} & \textbf{77.8} & \textbf{79.7} & \textbf{25} \\

\shline
\end{tabular}
\footnotesize
Overall best results are in bold.  
We reimplement FS-Net as \textbf{FS-Net(Ours)} for fair comparison because FS-Net uses different detection results.
\end{threeparttable}

\label{tab_real275}
\end{table*}

%% file: table/tab_ablation.tex
\begin{table*}[t]
\centering
\scalebox{1}{
\begin{threeparttable}
\caption{Ablation studies on NOCS-REAL275 datasets.}
    \centering
    \begin{tabular}{c|cc|c|c|c|cc|cccc}
    \shline
        ~ & Direct & Prior & Sym. $\mathcal{L}$ & Sym. Recon. & Removal & $IoU_{50}$ & $IoU_{75}$ &  $5^{\circ}2cm$& $5^{\circ}5cm$& $10^{\circ}5cm$& $10^{\circ}10cm$  \\ \hline
        A1 & \checkmark & ~ & ~ & ~ & ~ & 82.9 & 62.3 & 26.9 & 34.6 & 72.7 & 74.3  \\ 
        A2 & ~ & \checkmark & ~ & ~ & ~ & 81.1 & 48.3 & 10.6 & 19.8  & 55.9 & 59.3 \\       A3 & \checkmark & \checkmark & ~ & ~ & ~ & \textbf{83.3} & 63.7 & 27.5 & 38.3 & 72.0 & 73.6  \\\hline
        B1 & ~ & \checkmark & \checkmark & ~ & ~ & 83.2 & 58.7 & 30.5 & 41.0  & 71.2 & 73.3  \\ 
        B2 & \checkmark & \checkmark & \checkmark & ~ & ~ & 78.3 & 64.0 & 31.1 & 40.5 & 74.7 & 76.6  \\ \hline
        C & \checkmark & \checkmark & \checkmark & \checkmark & ~ & 83.1 & 66.0 & 33.3 & 42.1 & 76.1 & 78.0  \\ 
        \hline
        D & \checkmark & \checkmark & \checkmark & \checkmark & \checkmark & 82.3 & \textbf{66.3} & \textbf{34.7} & \textbf{44.6} & \textbf{77.8} & \textbf{79.7}  \\ \shline
    \end{tabular}
\footnotesize

\textbf{Direct} means using direct regression for pose, otherwise the pose is solved from predicted NOCS coordinates by Umeyama’s algorithm. 
\textbf{Prior} means using shape priors as input and performing shape prior deformation. 
\textbf{Sym. $\mathcal{L}$} includes the symmetry-aware loss terms for NOCS coordinate prediction and shape  deformation, otherwise the traditional loss terms are used. 
\textbf{Sym. Recon.} stands for symmetry reconstruction and \textbf{Removal} is the outlier removal process. 

\label{tab-abs}
\end{threeparttable}
}
\end{table*}

%% file: sections/experiment.tex
\section{EXPERIMENTS}

\textbf{Datasets.}
We use NOCS-REAL275 and NOCS-CAMERA25~\cite{NOCS} to evaluate SSP-Pose.
NOCS-REAL275 is a real-world dataset containing 4.3k images captured from 7 scenes for training and 2.75k images from 6 scenes for testing. It covers objects of 6 known categories, including \textit{bottle, bowl, camera, can, laptop} and \textit{mug}.
CAMERA25 is a synthetic dataset which is generated by rendering virtual object on the real background and shares the same categories with NOCS-REAL275. 
It provides 275k images for training and 25k for testing.

\textbf{Implementation Details.}
We follow~\cite{shape_deform,sgpa} to generate segmentation masks with an off-the-shelf object detector, \ie Mask-RCNN~\cite{maskrcnn} for fair comparison.
We employ data augmentation during training including random uniform noise and random rotational and translational perturbations similar to~\cite{fs-net}.
All the experiments are conducted on a single NVIDIA A100 GPU and a AMD EPYC 7H12 64-Core CPU.
The hyper parameters for all loss terms are kept unchanged during experimentation unless specified, with $\{\lambda_1, \lambda_2, \lambda_3, \lambda_4, \lambda_5\} = \{8.0, 10.0, 1.0, 1.0, 1.0\}$. 
We adopt the Ranger optimizer~\cite{ranger1,ranger2,ranger3} and set the batch size to be 32 and the base learning rate to be 1e-4.
The learning rate is annealed at $72\%$ of the training phase using a cosine schedule.
We only use the real data during training in the evaluation of dataset NOCS-REAL275.
We train a single model for all categories. 
The total training epoch is 150.

\textbf{Evaluation Metrics.}
We follow~\cite{NOCS, sgpa, dualposenet} to adopt 3D IOU that reports the mean precision of 3D intersection over union (IoU) at thresholds of 25$\%$, 50$\%$, 75$\%$ to jointly evaluate rotation, translation and size.
To directly compare errors in rotation and translation, we also adopt the $5^{\circ}2cm$, $5^{\circ}5cm$, $10^{\circ}5cm$, $10^{\circ}10cm$ metrics. 
A pose is thereby considered correct if the translation and rotation errors are both below the given thresholds.

\subsection{Comparison with State-of-the-Art Methods}
In table~\ref{tab_real275}, we compare SSP-Pose with state-of-the-art methods on NOCS-REAL275 dataset. 
Our method is superior to other methods under $IoU_{50}$, $IoU_{75}$, $5^{\circ}5cm$,  $10^{\circ}5cm$, $10^{\circ}10cm$ metrics, including all shape prior based methods and direct regression methods. 
In term of $IoU_{75}$, $5^{\circ}5cm$ and $10^{\circ}5cm$, our method surpasses the second best method by a large margin. Specifically, we achieve 82.5\% mAP under $IoU_{50}$, 64.2\% mAP under $IoU_{75}$, 44.5\% mAP under $5^{\circ}5cm$ and 75.7\% mAP under $10^{\circ}5cm$ respectively, where our performance is 2.4\%, 2.3\%, 4.9\% and 5.0\% better than the best shape prior based method SGPA\cite{sgpa}, and 2.1\%, 2.0\%, 8.6\% and 8.9\% better than the best direct regression method DualPoseNet~\cite{dualposenet}.

In table~\ref{tab_camera25}, we compare our method  with state-of-the-art methods On NOCS-CAMERA25 dataset. In $IoU_{75}$, $5^{\circ}5cm$ and $10^{\circ}5cm$ metrics, our result is superior to DualPoseNet and comparable to CR-Net and SGPA. 
Although CR-Net~\cite{cr-net} achieves better performance under $5^{\circ}2cm$ on the synthetic dataset CAMERA25, its accuracy on real dataset REAL275 is inferior in comparison with ours. Similarly, the margin in $5^{\circ}2cm$ between SPGA and ours is much smaller in the real world dataset.
The reason is that the noise in synthetic depth map is relatively small and does not contain outliers as discussed in Sec.~\ref{sec:filter}, which does not reflect the real-world situation. 

Moreover, our method is superior in running speed since only the pose regression head is needed during inference, as shown in the last column of Tab.~\ref{tab_real275}.
Regardless of the object detection time, the 9DoF pose estimation process can run at $>$50 FPS, which enables real-world applications with time requirements.
The framerate of our method reaches 25 FPS when using YOLOv3~\cite{tekin18_yolo6d} and ATSA~\cite{atsa} to segment the objects of interest. 

\begin{figure}[t]
  \centering
  \includegraphics[width=0.99\linewidth]{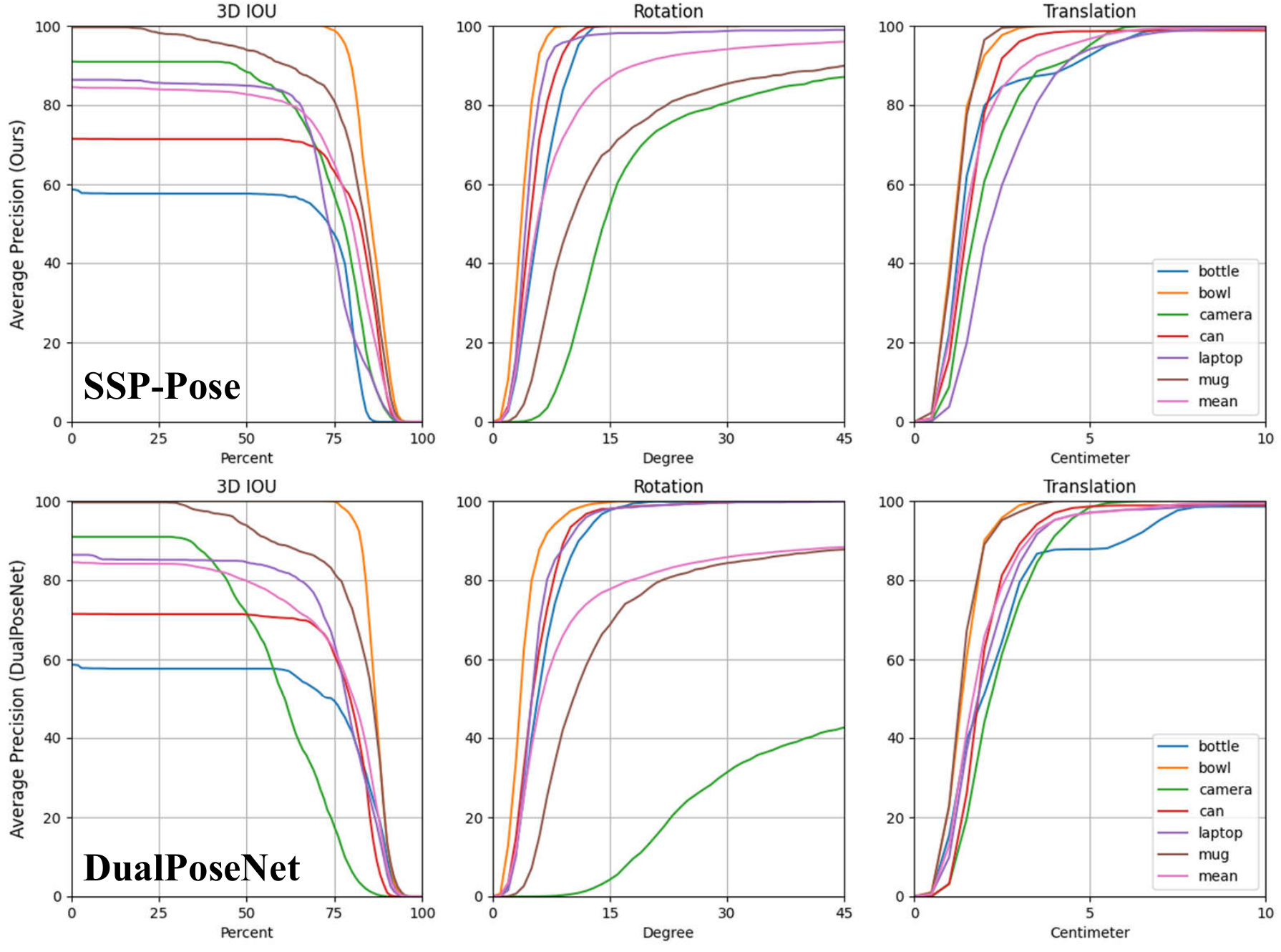}
  \caption{\textbf{Quantitative comparison of SSP-Pose and DualPoseNet\cite{dualposenet}.} We illustrate the precision under different error thresholds on NOCS-REAL275, including 3D IoU, rotation and translation. 
   }
  \label{quan-dpn}
\end{figure}

\begin{figure}[t]
  \centering
  ~\\[3pt]
  \includegraphics[width=0.99\linewidth]{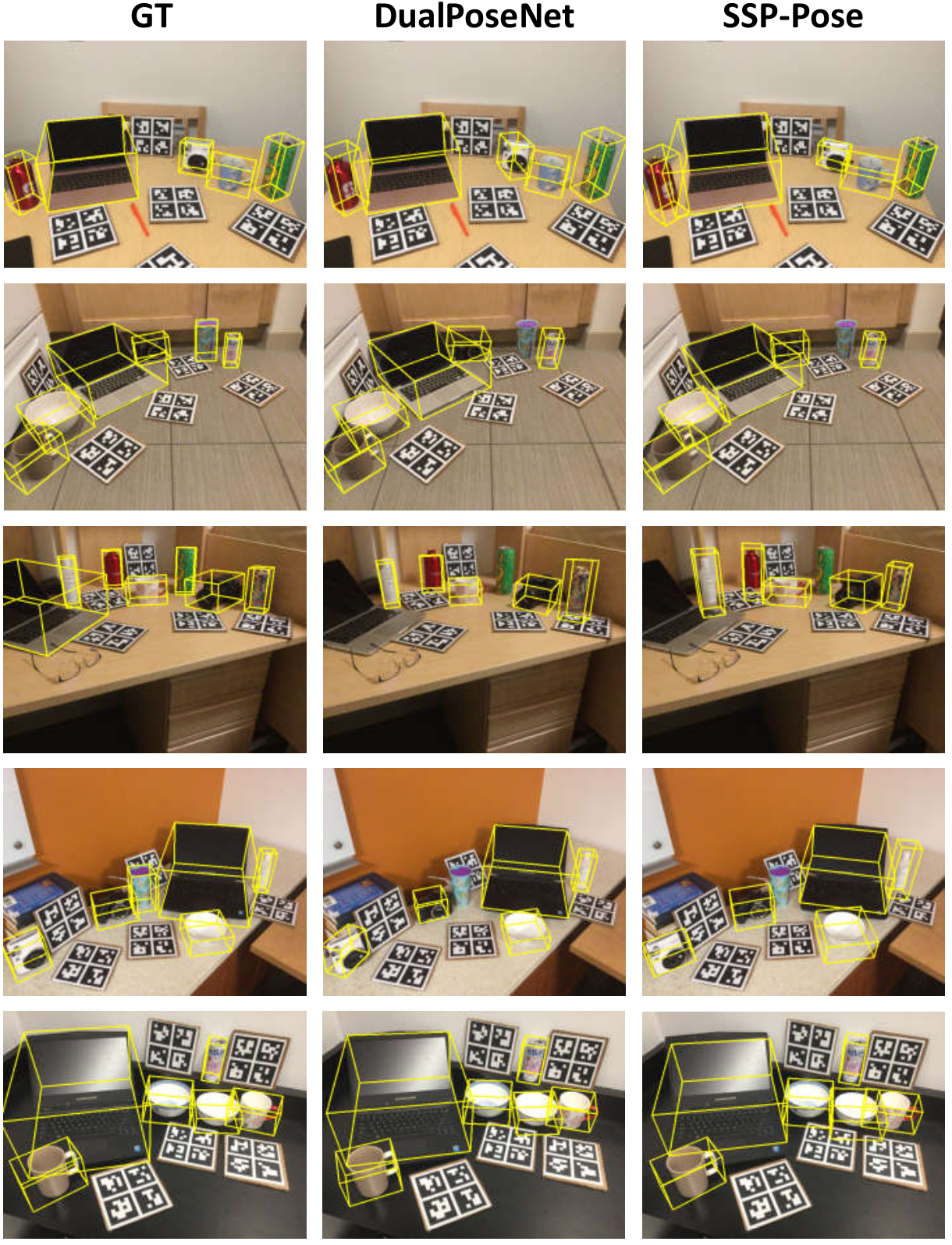}
  \caption{\textbf{Qualitative comparison of SSP-Pose and DualPoseNet\cite{dualposenet}.} 
   }
  \label{qual-dpn}
\end{figure}

We provide a detailed comparison with DualPoseNet~\cite{dualposenet} on REAL275 dataset in Fig~\ref{quan-dpn}.  
Our method produces significantly superior results in the accuracy of rotation and 3D IoU, especially for the category \textit{camera}.
This is also verified by the qualitative comparison in Fig~\ref{qual-dpn}. 
With the guidance of shape priors, SSP-Pose consistently outperforms DualPoseNet in handling complex objects such as \textit{camera}

\input{table/tab_camera}

\subsection{Ablation Studies}
\textbf{Effect of shape prior deformation and direct regression.}
In Tab.~\ref{tab-abs} (A) we conduct evaluation of the effect of shape prior deformation and direct regression. As can be observed, combining direct regression and shape prior deformation (A3) is superior to solely using each part (A1,A2) in all metrics.
Specifically, comparing A1 and A3, integrating category-level prior information and shape prior deformation process improves accuracy of $IoU_{75}$and $5^{\circ}5cm$ by 1.4\% and 3.7\% respectively.
From A2 to A3, by using direct regression for pose estimation instead of Umeyama's algorithm, the accuracy of $IoU_{75}$, $5^{\circ}2cm$, $5^{\circ}5cm$ and $10^{\circ}5cm$ improves more than 15\%, which indicates the effectiveness and robustness of direct pose regression.

\textbf{Effect of symmetry-aware loss.}
We evaluate the effect of the symmetry-aware loss in Tab.~\ref{tab-abs} (B). 
By comparing B2 and A3, the mAP under $5^{\circ}2cm$ and $5^{\circ}5cm$ improves 3.6\% and 2.2\% respectively, which proves that mitigating ambiguity caused by symmetry and utilizing symmetry prior bring significant improvement to pose estimation.
Interestingly, when recovering the pose using the Umeyama's algorithm (B1 and A2), using the symmetry-aware loss improves accuracy of $IoU_{75}$ and $5^{\circ}5cm$ by 10.4\% and 19.9\%, indicating that our symmetry-aware coordinate loss boosts the accuracy of coordinate prediction by a large margin.

\textbf{Effect of symmetry reconstruction.}
In Tab.~\ref{tab-abs} (C), we add symmetry reconstruction process to B1. The symmetry reconstruction improves $IoU_{75}$, $5^{\circ}2cm$ and $5^{\circ}5cm$ by 2.0\%, 2.2\% and 1.6\% respectively. 
The main reason is that symmetry reconstruction guides the network to learn the symmetry-aware geometric characteristics of each category for pose estimation.

\textbf{Effect of outlier removal.}
We verify the effectiveness of the outlier removal step in comparison of C and D. 
By only supervising point-wise loss terms on inliers and predicting point-wise mask, the accuracy of $5^{\circ}2cm$ and $5^{\circ}5cm$ improves by 1.4\%, 2.5\%, which illustrates the effectiveness of outlier removal technique.
Notably, this technique could be easily adopted for other methods in a plug-and-play manner.

%% file: table/tab_camera.tex
\begin{table}[t]
\centering
~\\[-5pt]
\caption{{Comparison with state-of-the-art methods on NOCS-CAMERA25 dataset.}
}
\begin{threeparttable}

\begin{tabular}{c|ccccc}
\shline
Method & $IoU_{75}$ &  $5^{\circ}2cm$& $5^{\circ}5cm$& $10^{\circ}5cm$\\
\hline
NOCS~\cite{NOCS}  & 69.5 & 32.3 & 40.9  & 64.6 \\
DualPoseNet~\cite{dualposenet}& 86.4 & 64.7  & 77.2 & 84.7 \\
SPD~\cite{shape_deform} & 83.1 & 54.3 & 59.0  & 81.5 \\
CR-Net~\cite{cr-net}  & 88.0 & \textbf{72.0} & \textbf{76.4}  & 87.7 \\
SGPA~\cite{sgpa}  & \textbf{88.1} & 70.7 & 74.5 & \textbf{88.4} \\

Ours  & {86.8} & {64.7} & {75.5} & {87.4} \\
\shline
\end{tabular}
\footnotesize
Overall best results are in bold. 
\end{threeparttable}

\label{tab_camera25}
\end{table}

%% file: sections/conclusion.tex
\section{CONCLUSION}
We propose an end-to-end category-level pose estimation network SSP-Pose, which deforms and associates the shape prior model to the observed object to establish consistency with the directly regressed pose, boosting performance whilst keeping a fast inference speed.
We utilize the category-level symmetry prior to guide the learning of shape prior deformation and pose regression.
Extensive experiments on public datasets demonstrate that SSP-Pose surpasses all competitors whilst keeping a real-time inference speed at 25Hz.
In the future, we plan to extend SSP-Pose to more complex scenes, \eg traffic scenes and robotic manipulation.